# Implementing Probabilistic Reasoning

Matthew L. Ginsberg

The Logic Group
Knowledge Systems Laborarory
Department of Computer Science
Stanford University
Stanford, California 94305

**Abstract.** General problems in analyzing information in a probabilistic database are considered. The practical difficulties (and occasional advantages) of storing uncertain data, of using it in conventional forward- or backward-chaining inference engines, and of working with a probabilistic version of resolution are discussed. The background for this paper is the incorporation of uncertain reasoning facilities in MRS, a general-purpose expert system building tool.

## §1. Introduction

There has been a great deal of work in the past few years concerning the theoretical underpinnings of various methods of inexact reasoning, including, among others, MYCIN-type certainty factors [5], Zadeh's fuzzy sets [7] and Dempster-Shafer theory [1,4]. Inclusion of these ideas in practical systems seems to have lagged, however, with the possible exception of the appearance of EMYCIN, an expert-system-building tool using MYCIN's inference engine.

This is unsatisfactory for a variety of reasons. Most importantly, the true advantages of the various competing paradigms which have been developed will only be apparent when these paradigms have been incorporated in full-scale systems. Until that time, the merits of various specific schemes must remain only expectations.

In addition, the development of an implementation often serves to clarify theoretical points which might otherwise go unnoticed; this has certainly been the case with the author's experience in incorporating a simplified version of Dempster's rule into the MRS expert system tool.

It is in some sense a chronicle of this implementation that I would like to present in this paper. Ideally, a description of this work will both encourage the proponents of other approaches to undertake similar projects and make it easier for them to do so. It may be my personal view that fuzzy sets are computationally intractable and that the unavailability of prior probabilities makes Bayesianism too naïve; the aim of this paper is to help the defenders of these methods prove me wrong.

## §2. An overview of MRS

MRS ("Meta-level Representation System") is an expert system building tool currently used by approximately twenty AI development groups. As of July 1984, it was purely predicate-calculus based: True facts were simply stored in the MRS database.

Not surprisingly, the various MRS users have found this to be unsatisfactory for the development of systems operating in uncertain domains. Their solution has generally had to be to incorporate some sort of certainty factors into MRS by hand, producing rules such as

```
(if (and (cf premise $c) (> $c 0.6)) (cf conclusion $c)).
```

This amounts to saying that if the certainty factor assigned to the premise of some rule is greater than 0.6, then the conclusion should receive the same certainty factor as the premise. (MRS prefixes variables with $ signs.)

The key idea behind MRS is that of user-specified control. Suppose that the user wants to inform the system that to prove a statement of the form (foo $x), resolution should be used. He would do this by adding the fact

$$(\text{totruep '(foo \$x) resolution}) \tag{1}$$

to the MRS database. If the user now asked MRS to prove (foo fred) by typing (truep '(foo fred)), MRS would proceed by first proving

$$(\text{totruep '(foo fred) \$m}).$$

This succeeds, returning an answer with $m bound to resolution. MRS therefore proceeds by invoking the resolution theorem prover, evaluating (resolution '(foo fred)).

The inference methods available to the MRS user include simple lookup, forward and backward chaining, and resolution. This was one of the reasons that incorporating probabilistic reasoning facilities into the system was attractive from a research point of view — it would naturally provide information regarding the relative efficacy of uncertain reasoning within each of these techniques.

The appearance in MRS of meta-level rules such as (1) is another reason the incorporation of probabilities was of interest. The issue of control is an extremely complicated one; it may well be the case that control of inference in complex situations should proceed on a probabilistic basis. Early results in the run-time control of backward chaining seem to support this [6].

The subsequent sections of this paper will address the problems of dealing with the various MRS inference techniques probabilistically. Section three deals with the problem of simply storing and retrieving information from a probabilistic database. As there appear to be significant advantages to doing this using a belief-disbelief approach such as that in Dempster-Shafer theory, this section will be directed substantially toward this sort of implementation.

The remaining sections will deal with probabilistic inference more generally, and will discuss forward chaining, backward chaining and resolution in turn. The final section will summarize some of the apparent practical advantages and difficulties of probabilistic inference.

## §3. Probabilistic databases

The idea here is a simple one: Instead of simply storing facts in a database, pairs are stored, where the first element of the pair is the fact to be stored and the second is the probabilistic truth value.

In the MRS implementation, the truth values consist of pairs (a . b) where $a$ corresponds to the extent to which the available evidence confirms the fact and $b$ to the extent to which it disconfirms it. (As a consequence, $a + b \leq 1$ for all facts in the database.) True, or totally confirmed facts, have truth value (1 . 0). False, or totally disconfirmed facts, have truth value (0 . 1). Thus, to assert (not (foo fred)), we could stash into the database *either* of the pairs

$$( \text{(not (foo fred)) (1 . 0)} )$$

or

$$( \text{(foo fred) (0 . 1)})$$

This was deemed to be unacceptable. One of the principal advantages of the Dempster-Shafer approach is that it enables one to combine information about both the confirmation and disconfirmation of a single hypothesis; it therefore seemed natural to choose the second of the above expressions. In general, when the user attempts to stash (not $x) in the database with truth value (a . b), the action taken by MRS is to stash $x with truth value (b . a).

A similar modification was required when information was retrieved from the database. A lookup of (foo fred) is interpreted as a request for the confirmation of (foo fred), and the system therefore returns simply the confirmation in the associated truth value. Since the confirmation of (not (foo fred)) is the same as the *disconfirmation* of (foo fred), the value returned from looking up (not (foo fred)) is the disconfirmation of the truth value associated with (foo fred).

There proved to be other useful results that could be returned from a given truth value. In looking up (unknown (foo fred)) for example, it seems natural to return $1 - a - b$ where $a$ is the confirmation and $b$ the disconfirmation of (foo fred).

A function which converts a truth value (a . b) to a single number will be referred to as a *tag*. In addition to the three already discussed,

$$t : (a \ . \ b) \rightarrow a$$
$$not : (a \ . \ b) \rightarrow b$$
$$unknown : (a \ . \ b) \rightarrow 1 - a - b$$

there are also

$$poss : (a \ . \ b) \rightarrow 1 - b$$
$$poss\text{-}not : (a \ . \ b) \rightarrow 1 - a$$
$$mass : (a \ . \ b) \rightarrow a + b.$$

Poss measures the extent to which a given statement is possible (i.e., not disconfirmed), and poss-not the extent to which it is possibly false. Mass measures the total extent to which information about the statement is available in the database.

In the implementation, there is a single function lookup which accepts three arguments: a sentence such as (foo $x), a tag (which defaults to t) and a cutoff (which defaults to 1.0). If a fact matching the supplied one can be found in the database such that the result of applying the supplied tag to the truth value of the database fact is no less than the supplied cutoff, an appropriate binding list is returned. Thus, if in our database we had

$$(foo\ fred) \quad (0.3 \ . \ 0.2)$$
$$(foo\ harry) \quad (0.7 \ . \ 0.0)$$

the result of (lookup (foo $x) t 0.5) would be a list binding $x to harry, since only (foo harry) has confirmation of 0.5 or greater.

## §4. Forward chaining

In conventional forward chaining, when a new fact is added to the database, its consequences are also added. It is generally assumed that each of these consequences is consistent with the other information in the database, so forward chaining is simply a matter of applying each of the rules in the database to the new piece of information and its consequences.

This is not satisfactory probabilistically. The reason is that it is very commonly the case that the information being added to the database is inconsistent with knowledge already there. Typically, some statement such as (foo fred) will be stored in the database with truth value (a . b), and the new information being passed to the forward chainer is that the truth value should in fact be (c . d). Not only will the truth value assigned to (foo fred) need to be changed, but the truth values of its consequences will need to be modified as well. The problem persists if some other sort of probabilistic representation is used.

In order that we have a concrete example, suppose that we have in our database the rule

$$(if\ (foo\ \$x)\ (goo\ \$x)). \tag{2}$$

If we now replace the truth value of (foo fred) with (c . d), it follows that there will be a new contribution of (c . d) to the truth value of (goo fred). (Again, the points I am about to make hold for other probabilistic inference schemes as well; I am using the confirmation/disconfirmation representation only for definiteness.)

A simplistic approach at this point would be to simply replace the truth value of (goo fred) with (c . d), but this does not account for the fact that the truth value of (goo fred) may in fact be the result of combining contributions from many different sources. In actuality, if the rule (2) has been applied using (foo fred)'s previous truth value (a . b), only the contribution to the truth value of (goo fred) generated by this application needs to be retracted when the forward chainer is invoked.

There are two implications to this. The first is that the rule of combination being used in a probabilistic inference scheme must be invertible. In the example we are considering, if the truth value of (goo fred) is an accumulation from various sources, it must be possible to remove one term in this accumulation without affecting the others. The combining rule used in MRS is that which I presented at AAAI-84 [2]; I noted there that the rule was invertible, and described the inverse to it.

The second point to be made is that probabilistic inference schemes must be equipped with some minimal sort of reason maintenance facility. We noted above that the result of applying (2) to (foo fred) needed to be inverted *if (2) had been applied to (foo fred)'s previous truth value*. It follows that the forward chainer needs access to a list of the rules in the database which have already been applied, together with the truth values of the antecedents at the time of application.

There are also representational issues that need to be addressed in the forward chainer and in probabilistic inference generally. The rule (2) above states that (foo $x) implies (goo $x). Suppose, however, that we wanted the rule to state that (foo $x) implied (not (goo $x)). One solution would be to stash (2) in our database with truth value (0 . 1) (i.e., false). The difficulty with this is that

$$(\text{not (if (foo \$x) (goo \$x))}) \tag{3}$$

is not logically equivalent to

$$(\text{if (foo \$x) (not (goo \$x))}). \tag{4}$$

It seems in fact that probabilistic rules must in fact be of the form

$$(\text{if premise consequence truth-value}), \tag{5}$$

where truth-value is the truth value to be assigned to the consequence if the premise is true. How to propagate partial truths through a rule such as (5) is a theoretical question which will need to be addressed by the probabilistic theory being considered. The solution used in the MRS implementation is described in [2].

The rule (4) now can be represented as

$$(\text{if (foo \$x) (goo \$x) (0 . 1)}),$$

while the statement (3) can be recorded by storing

$$(\text{if (foo \$x) (goo \$x) (1 . 0)})$$

in the database with truth value (0 . 1).

The appearance of the extra truth value in (5) can also be understood in terms of a difference between the conditional probability $p(A|B)$ and the probability of a conditional $p(A \to B)$.

Instances of $\neg A$ will increase the probability $p(A \to B)$ while having no effect on $p(A|B)$; any inference scheme which generalizes predicate calculus needs to preserve this distinction.

It is also interesting to note that because the value in (5) is that of the *conditional probability*, it will not be affected by negative instances of the premise: thus a non-black non-crow cannot be interpreted as confirmation for the hypothesis that all crows are black, since there is no useful relationship between the conditional probabilities $p(A|B)$ and $p(\neg B|\neg A)$.

### 4.1 Control of forward chaining

Finally, suppose that instead of (2) we had

```
(if (foo $x) (goo $x) (0.01 . 0.0)),
```

so that the truth of (foo $x) increased the confirmation of (goo $x) very slightly. If the truth value of (foo $x) changes only marginally, it may well be the case that the corresponding increment to the truth value of (goo $x) is so small that we do not wish to consider consequences of it. A convenient way to implement this is to take advantage of the reason maintenance facilities described earlier: If the truth value of the premise of a rule changes only very little, the previous result is not retracted, and the rule is not re-fired. MRS uses a variable inference-cutoff; if the mass of the difference between the previous and current truth values of the premise of a rule is less than inference-cutoff, the forward chainer takes no action when the rule is encountered.

## §5. Backward chaining

Conventional logical backward chainers proceed very simply: Given a fact to be proved, they first search the database for the fact. If it is not found, they find a rule whose consequent matches the fact. Having found one, the backward chainer recursively tries to prove the premise of the rule. Success at any point represents a proof of the original assertion.

The probabilistic case is complicated by the fact that truth values accumulate from a variety of sources. Suppose we are trying to prove (flies Tweety) for some bird Tweety, and have rules in our database

```
(if (bird $x) (flies $x) (0.7 . 0.0))
(if (ostrich $x) (flies $x) (0 . 1))
```

In proving (flies Tweety), it does not suffice to apply the first of these and succeed — the second must also be considered. In this simple case, we must save the (accumulated) truth value of (flies Tweety) and continue with the derivation until there are no more possible rules affecting the truth value of the supplied conclusion.

The situation is complicated further by the fact that the statement being proved may not be grounded. MRS interprets an attempt to prove (flies $x) as an attempt to find a binding for $x which make (flies $x) true; in proving such a statement, it is necessary to store a list of bindings for the variables in the original statement with associated truth values accumulated for each.

In fact, this problem can occur even if the proposition being proved *is* grounded. Consider the rule:

```
(if (steals $person $object) (crook $person)).
```

If we try to prove the ground assertion (crook Nixon), the recursive definition of backward chaining will result in our attempting to prove the non-grounded statement (steals Nixon $object).

The problem of accumulating truth values is especially difficult if the backward chainer is agenda-based. (As the MRS backward chainer is, to allow for maximum use of meta-level control information.) The reason for this is that many tasks on the agenda will need to access the same list of bindings and partial truth values, and it is quite possible that the list used by some task $t$ will be modified by other probabilistic inferences between the time $t$ is added to the agenda and the time $t$ is actually executed.

### 5.1 Controlling backward inference

As in forward chaining, it may be desirable to cut off a backward inference if its total effect on the truth value of the statement being proved is small. If we had the rule

```
(if (politician $p) (crook $p) (0.1 . 0.0))
```

it might well be that proving someone to be a politician would have an effect small enough on the truth value of his being a crook as to be negligible—in the MRS implementation, the inference is again terminated if the mass of the eventual contribution can be shown to be less than inference-cutoff. A similar facility appears in EMYCIN, which does not consider the application of rules of inference which will affect the certainty factor associated to a given conclusion by an amount of 0.2 or less.

Another way in which backward inference can be terminated early is hinted at in the bird/ostrich example above. In situations where the speed of inference is critical, it may be desirable to simply accept a statement with a confirmation of 0.9 (say) as true, and not to expend additional effort in trying to prove it false. This is implemented in MRS through the variable accept-as-true; if the confirmation or disconfirmation of a particular statement is greater than or equal to this value, the staement is accepted as confirmed or disconfirmed even though subsequent analysis might conceivably overturn this conclusion.

## §6. Resolution

Resolution has proven to be the most difficult of the MRS inference methods to implement probabilistically. In addition to the reappearance of the practical difficulties described for backward chaining, there are also significant theoretical issues to be resolved. I discuss these elsewhere [3]; let me confine myself here to a few comments about the nature of the difficulty.

The basic resolution rule of inference is a consequence of the logical implication

$$(p \vee q) \wedge (\neg p \vee r) \to (q \vee r). \tag{6}$$

Probabilistically, the truth value of $p \vee q$ also contains information about the truth of $\neg(p \vee q)$; it is therefore possible to "resolve" $(p \vee q)$ with $(p \vee r)$ using the implications

$$\neg(p \vee q) \wedge \neg(p \vee r) \to \neg(q \vee r) \tag{7}$$

and

$$[(p \vee q) \wedge \neg(p \vee r)] \vee [\neg(p \vee q) \wedge (p \vee r)] \to (q \vee r). \tag{8}$$

A probabilistic resolution theorem prover should incorporate inferences made possible by these implications.

An additional difficulty arises because of the nature of the implications (6)–(8) themselves. In evaluating the truth value to be associated to a conjunction, the individual conjuncts are often assumed to be independent; in an expression such as the one appearing on the lefthand side of (6), this is explicitly not the case. The truth value associated to the premise of this rule must therefore be evaluated bearing this in mind. Again, details can be found in [3].

## §7. Conclusion

Relative to my expectations when I began this work, there were two surprising sources of difficulty and two unforeseen advantages to implementing non-monotonic reasoning probabilistically. Let me summarize the advantages first.

The first has to do with the possibility of terminating probabilistic inference early if either the conclusion becomes extremely likely or the contribution resulting from the inference under consideration will be small. Although these procedures are not non-monotonically sound, they allow a practical implementation to avoid the problems that would otherwise arise due to the fact that non-monotonic inference is fundamentally undecidable.

The second advantage is the uniform treatment of negation allowed by a probabilistic scheme. The previous version of MRS stored (foo fred) and (not (foo fred)) separately in the database; a probabilistic scheme makes clear the connection between the two.

Surprisingly, this uniform treatment of negation was the source of many of the difficulties encountered. The need to rewrite implications such as

```
(if (foo $x) (not (goo $x)))
```

caused a considerable amount of difficulty; the treatment of negation also led to theoretical problems in dealing with resolution.

The most serious difficulty encountered, however, has proven to be the need to keep a list of "partial answers" in any backward-directed inference procedure (either backward chaining or resolution). As mentioned earlier, the fact that the chosen implementation needed to be agenda-based only compounded this problem.

The probabilistic version of MRS is scheduled to be released publicly later in 1985. The author is looking forward both to its reception by the user community, and to the possibility of comparing the performance of Dempster-Shafer methods with implementations based on different theories.

## Acknowledgement

This research has been supported by GTE/Sylvania and by the Office of Naval Research under grant number N00014-81-K-0004.